\documentclass[10pt,twocolumn,letterpaper]{article}

\usepackage[pagenumbers]{cvpr} 

\usepackage{graphicx}
\usepackage{amsmath}
\usepackage{amssymb}
\usepackage{booktabs}
\usepackage{multirow}
\usepackage{lipsum}

\usepackage[dvipsnames]{xcolor}
\usepackage[pagebackref,breaklinks,colorlinks]{hyperref}

\title{Direct Object-Level Reconstruction via Probabilistic Gaussian Splatting}

\author{
Shuai Guo\textsuperscript{1,2} \quad
Ao Guo\textsuperscript{1,2} \quad
Junchao Zhao\textsuperscript{1,2} \quad
Qi Chen\textsuperscript{1,2} \quad
Yuxiang Qi\textsuperscript{3} \\
Zechuan Li\textsuperscript{1,2} \quad
Dong Chen\textsuperscript{1,2} \quad
Mingliang Xu\textsuperscript{1,2\dag} \\
\textsuperscript{1}School of Computer and Artificial Intelligence of Zhengzhou University, 450001, China \\
\textsuperscript{2}Zhengzhou University Engineering Research Center of Intelligent Swarm Systems, 450001, China \\
\textsuperscript{3}School of Data Science and Artificial Intelligence, \\Dongbei University of Finance and Economics, Dalian, 116025, China \\
}

\begin{document}
\maketitle

\let\thefootnote\relax\footnotetext{
\textsuperscript{\dag}Mingliang Xu is the corresponding author. 
This work was supported in part by the National Natural Science Foundation of China under Grant No. 61903341 and 62325602; in part by the Key Scientific Research Project Plan of Colleges and Universities in Henan Province under Grant No. 26A520039, in part by the Foundation of the State Key Laboratory of Robotics of China under Grant No. 2022-KF-22-06; in part by the National Key Research and Development Program Project under Grant No. 2024YFB3311600.
}

\begin{abstract}
Object-level 3D reconstruction play important roles across domains such as cultural heritage digitization, industrial manufacturing, and virtual reality. However, existing Gaussian Splatting–based approaches generally rely on full-scene reconstruction, in which substantial redundant background information is introduced, leading to increased computational and storage overhead. To address this limitation, we propose an efficient single-object 3D reconstruction method based on 2D Gaussian Splatting. By directly integrating foreground–background probability cues into Gaussian primitives and dynamically pruning low-probability Gaussians during training, the proposed method fundamentally focuses on an object of interest and improves the memory and computational efficiency. Our pipeline leverages probability masks generated by YOLO and SAM to supervise probabilistic Gaussian attributes, replacing binary masks with continuous probability values to mitigate boundary ambiguity. Additionally, we propose a dual-stage filtering strategy for training's startup to suppress background Gaussians. And, during training, rendered probability masks are conversely employed to refine supervision and enhance boundary consistency across views. Experiments conducted on the MIP-360, T\&T, and NVOS datasets demonstrate that our method exhibits strong self-correction capability in the presence of mask errors and achieves reconstruction quality comparable to standard 3DGS approaches, while requiring only approximately 1/10 of their Gaussian amount. These results validate the efficiency and robustness of our method for single-object reconstruction and highlight its potential for applications requiring both high fidelity and computational efficiency.
\end{abstract}

\vspace{0.5em} \noindent \textbf{Keywords:} 2D Gaussian Splatting, Single-Object Reconstruction, Probabilistic Supervision, Background Filtering

\section{Introduction}
Object-level 3D reconstruction is highly valuable but has not been sufficiently explored in the field of computer vision. The emergence of 3D Gaussian Splatting (3DGS) \cite{Kerbl2023} offers new opportunities to advance research in this domain. 3DGS employs a dense set of Gaussian primitives to form an explicit scene representation, which improves both training efficiency and reconstruction quality through machine learning. Furthermore, 2D Gaussian Splatting (2DGS) \cite{Huang2024} addresses the view-inconsistency problem inherent in 3DGS, representing another major advancement in Gaussian Splatting-based reconstruction methods.

However, 2DGS, 3DGS, and most subsequent research built upon them primarily focus on full-scene reconstruction, whereas many real-world applications require reconstructing only a specific object. For example, the digital preservation of cultural relics or the geometric modeling of industrial objects typically focuses on specific objects of interest, while regarding the environment as unnecessary background, which incurs significant additional storage and computational costs in full-scene reconstruction. Consequently, for these applications, direct single-object reconstruction has been a fundamental requirement. A fundamental challenge in direct single-object reconstruction is to robustly separate the object from the background, a step which is critical for accurate reconstruction. In this study, we propose an efficient method for single-object reconstruction based on 2DGS. Unlike the conventional “\textit{full-scene-reconstruct then object-segmentation}” paradigm, which applies semantic segmentation only after full-scene modeling, our method directly incorporates segmentation cues into Gaussian attributes and dynamically suppresses or removes background Gaussians during training. This design effectively reduces the storage and computational overhead associated with redundant background modeling while preserving the fine-grained geometric and appearance details of the target object. Furthermore, to reduce the number of Gaussians, we introduce a probability-mask-based filtering strategy that removes redundant background points during initialization, thereby preventing the generation of unnecessary background Gaussians and substantially improving overall reconstruction efficiency.

Our proposed method for single-object reconstruction augments each Gaussian primitive with a probabilistic attribute to determine whether it belongs to the target object, as a result, we could render synthesized masks for all views. However, effectively supervising these probabilistic attributes during training remains challenging. To address this, we first preprocess the RGB images from multiple views and use You Only Look Once (YOLO) \cite{Varghese2024} and Segment Anything Model (SAM) \cite{Kirillov2023} to generate pixel-wise probability masks, where each pixel value indicates its likelihood of belonging to the target object. These probability masks, together with the synthesized masks, support the supervised learning of the probabilistic attributes of Gaussian primitives. In summary, our contributions are summarized as follows.
\begin{itemize}
\item We propose an efficient method for direct object-level reconstruction. Our pipeline substantially reduces the memory and computational overhead during training, the peak number of Gaussian primitives is only one-tenth of that needed for full-scene reconstruction.
\item We propose a two-stage strategy to integrate 2D segmentation. The first stage suppresses background interference using SAM-predicted masks, while the second stage refines object boundaries via rendering-aware segmentation.
\item We further introduce an initialization refinement strategy that effectively removes points from background while preserving the majority points belonging to the target object, thereby substantially reducing the number of initialized Gaussians.
\end{itemize}

\section{Related Work}

\subsection{Model Compression in 3D Gaussian Splatting}
Multi-view reconstruction is a fundamental task in 3D vision. Recently, 3DGS \cite{Kerbl2023} has demonstrated efficient training and real-time rendering by explicitly representing discrete Gaussian primitives in 3D space, achieving high visual fidelity in reconstruction \cite{Fei2024}. However, the application of 3DGS in large-scale open scenes still faces significant challenges. Specifically, fitting an entire scene often necessitates introducing a huge amount of Gaussian primitives, resulting in substantial storage and computational overhead. To address this issue, various model compression approaches have recently been proposed to reduce the resource consumption of 3DGS in practical applications. Conceptually, these approaches can be divided into model compression and model compaction. The core idea of model compression is to reduce redundant Gaussian attributes or optimize the storage structure, thereby lowering memory overhead. Common implementations include vector quantization, attribute pruning, and structured optimization. For example, \cite{Navaneet2023,Fan2024,Niedermayr2024} reduce memory usage by vectorizing similar Gaussian attributes, thereby eliminating redundant storage. Approaches in \cite{Lu2024,Wang2024} partition Gaussian distributions using anchor points, achieving efficient compression without strict spatial subdivision. Approaches in \cite{Lee2024,Papantonakis2024} compress color attributes through integration or adaptive adjustment, further reducing the overall model size. In contrast, model compaction focuses on optimizing the number and spatial distribution of Gaussian primitives, aiming to remove redundant primitives while preserving reconstruction quality. For example, approaches in \cite{Cheng2024,Liu2024} employ depth and normal images during training to provide geometric guidance, enabling more precise pruning and densification, thereby avoiding excessive Gaussian reconstruction and achieving model compaction. Meanwhile, \cite{Mallick2024} employs a global scoring strategy to guide the addition of Gaussian primitives, ensuring efficient densification. And, in \cite{Kim2024}, color gradients are incorporated during densification to further enhance guidance.

Moreover, since naive 3D Gaussian primitives exhibit an approximately ellipsoidal shape in space, they have inherent limitations in accurately representing surface geometry, often requiring dense and redundant Gaussian primitives to approximate the real-world surface. Further, in splatting-based rendering, the intersection-calculation with the viewing ray is typically approximated using the center of each Gaussian primitive rather than the true intersection point, which arises the view inconsistency \cite{sugar} and introduces misleading information in subsequent densification processes. To mitigate these issues, existing studies primarily flatten Gaussian primitives and incorporating geometric constraints during training to improve their ability to fit real surfaces \cite{sugar, Chen2024,Jiang2024,Zhang2024}. Building upon this idea, 2DGS \cite{Huang2024} further simplifies the traditional ellipsoidal Gaussian representation into a 2D disk, optimizing the splatting-based rendering process and effectively preventing view inconsistency. The 2D Gaussian primitives inherently encode normal information, which not only provides more reliable geometric guidance during pruning and densification, thereby preventing over-reconstruction, but also contributes to reducing the total number of Gaussian primitives.

However, the aforementioned approaches still primarily focus on full-scene reconstruction, which inevitably introduces a substantial amount of redundant background Gaussian primitives. Such redundancy not only incurs additional computational overhead but may also lead the model to overemphasize background details during training, particularly in open scenes. In contrast, our proposed method can directly exclude background regions from the reconstruction process. Consequently, the modeling process is explicitly focused on the target object, sharply reducing the number of Gaussian primitives required. Experimental results demonstrate that the peak number of Gaussians in our approach is only about one-tenth of that in conventional 3DGS for full-scene reconstruction, as effective background removal enables high-quality object reconstruction with far fewer primitives. Overall, our approach focuses on specific objects of interest and significantly reduces both memory and computational demands.

\subsection{Object Reconstruction via Segmentation}
The photorealistic and real-time rendering capabilities of 3DGS have been demonstrated in a wide range of 3D vision tasks. Prior segmentation approaches based on Neural Radiance Fields (NeRF) \cite{Mildenhall2021} have shown the feasibility of projecting 2D semantic information into 3D space, providing new avenues for semantic segmentation in 3D domain. For instance, \cite{Cen2023} leverages 2D segmentation cues to generate masks within the NeRF rendering process and projects them into 3D semantic representations via multi-view consistency. \cite{Chen2023} further integrates pre-trained features with interactive prompts to enable controllable 3D segmentation. In parallel, \cite{Wang2022} explores geometry decomposition guided by 2D images to facilitate structured operations in 3D scenes, while \cite{Fu2022} addresses label propagation from 2D to 3D to improve panoramic segmentation in large-scale scenes.

Inspired by the aforementioned approaches, recent studies have increasingly explored incorporating semantic segmentation information into the 3DGS reconstruction process to segment scenes and indirectly obtain object-level Gaussian models. So far, the studies for single-object 3DGS reconstruction can be broadly classified into two categories. The first involves performing full-scene 3DGS reconstruction as the initial step, followed by extraction and separation of the object using 2D semantic segmentation cues. Typically, these approaches rely on 2D segmentation models such as SAM \cite{Kirillov2023} to obtain semantic cues that supervise the semantic feature attributes of Gaussian primitives, thereby constructing Gaussian semantic fields for scene segmentation \cite{Lan2024,Guo2024,Choi2024}. Among these approaches, SAGA \cite{Cen2025a} estimates a semantic posterior probability for each Gaussian by aggregating multi-view 2D segmentation outputs, whereas Gaussian Group \cite{Ye2024} employs a compact identity encoding to classify Gaussian primitives. To further accelerate segmentation, FlashSplat \cite{Shen2024} reformulates the segmentation process as a linear regression problem, avoiding the time overhead associated with Gaussian optimization. Additionally, SAGD \cite{Hu2024} addresses boundary roughness by decomposing boundary Gaussians, assigning binary labels to each decomposed Gaussian primitive, and using a voting strategy to determine segmentation, substantially improving boundary precision. In contrast, the second category directly embeds semantic information during model training and performs Gaussian primitive segmentation based on the learned feature representations. For example, Feature3DGS \cite{Zhou2024} augments 3DGS Gaussian primitives with semantic feature attributes and supervises them during training. Contrastive Gaussian Clustering \cite{Silva2024} attaches a feature vector to each 3DGS Gaussian primitive and introduces a spatial similarity regularization loss during training, encouraging adjacent Gaussian primitives to have similar features while distant Gaussian primitives differ, thereby improving training performance. Similarly, COB-GS \cite{Zhang2025} jointly optimizes the 3DGS Gaussian model while incorporating segmentation information to guide volumetric optimization, resulting in more precise segmented Gaussian models.

It should be noted that both of the above categories follow the \textit{“full-scene reconstruction then object segmentation”} paradigm and share a common limitation: they inevitably require full-scene reconstruction prior to segmentation. In large-scale open scenes, excessive background information not only introduces redundant computational overhead but also significantly increases hardware resource demands. In comparison, our method directly reconstructs the target object, rather than extracting it from a pre-built full-scene reconstruction. Throughout the training process, our approach integrates foreground-background probability cues into the reconstruction pipeline and dynamically prunes low-probability background Gaussian primitives, resulting in a clean object-level Gaussian model. This focuses computational resources on the fine-grained reconstruction of the target object, maintains a low Gaussian count, and reduces both computational and memory overhead.

\section{Method}

\begin{figure*}[!htbp]
  \centering
  \includegraphics[width=\linewidth]{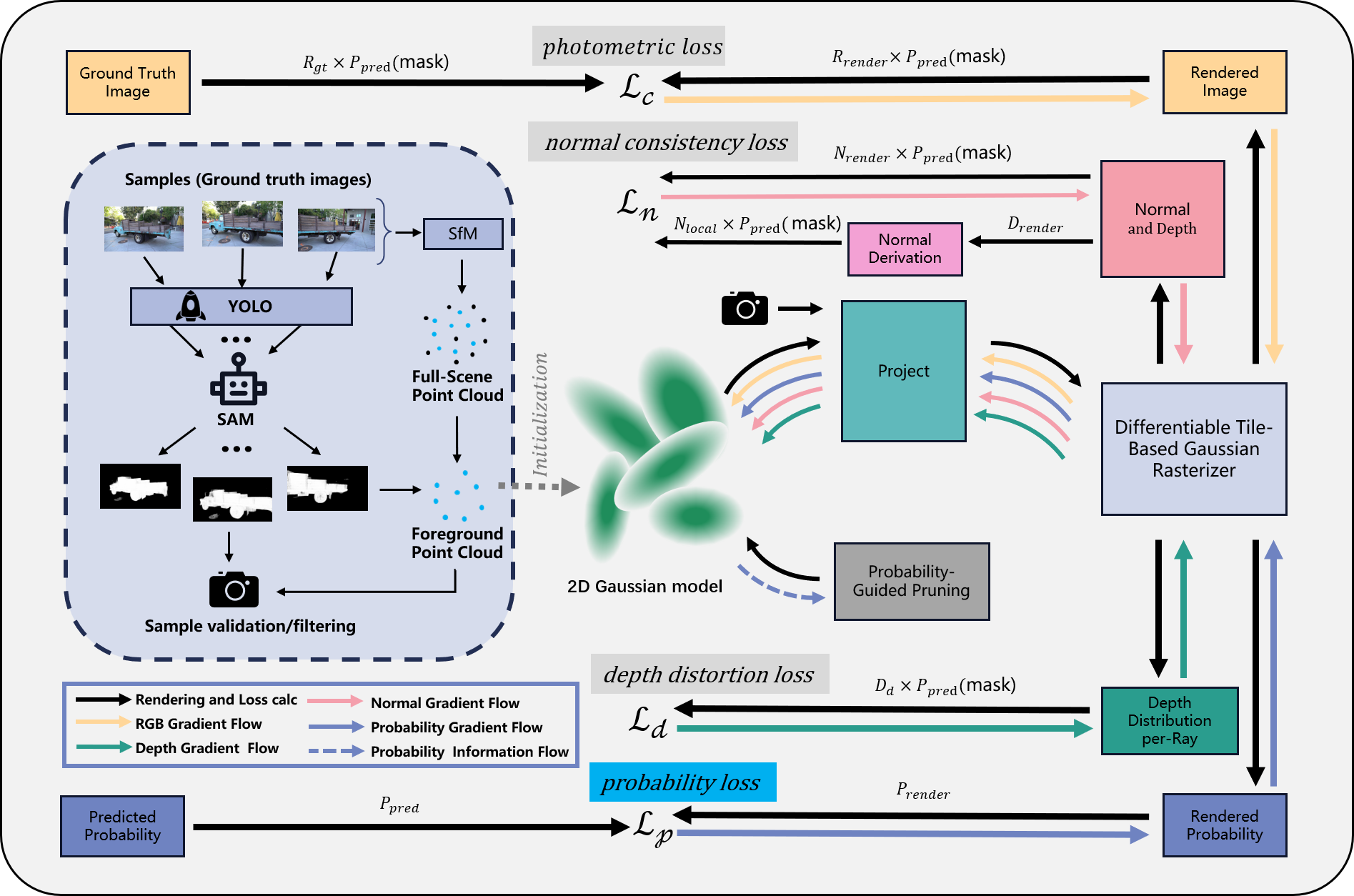}
  \caption{Overall pipeline of the proposed direct object-level reconstruction framework. The framework consists of two main components: data preprocessing on the left and Gaussian optimization on the right. In the preprocessing stage, YOLO and SAM are used to generate semantic probability masks for filtering the SfM point cloud and selecting valid views. During training, Gaussian primitives are initialized from the filtered point cloud and optimized by sampling valid views. A probability-based pruning module dynamically removes background Gaussians, enabling efficient and accurate single-object reconstruction.}
  \label{fig:fig1}
\end{figure*}

In this section, we introduce the reconstruction pipeline of our proposed method in three parts. First, Sec ~\ref{sec:data preparation} presents the data preparation process, including the acquisition of probability masks and the generation of the initial point cloud. Here, we deep analyze two critical issues inherent in the initial data: the background redundancy in initialization and mis-segmentation problem. To address these issues, Sec ~\ref{sec:data refinement} proposes a foreground point cloud extraction method along with a low-quality sample filtering strategy. Finally, Sec ~\ref{sec:Probabilistic Single-Object Gaussian Model and Optimization} provides a comprehensive description of the probabilistic Gaussian modeling approach and its corresponding optimization strategy. Fig.~\ref{fig:fig1} illustrates the overall pipeline of our proposed method.

\subsection{Data Preparation}
\label{sec:data preparation}

\begin{figure}
  \centering
  \includegraphics[width=\linewidth]{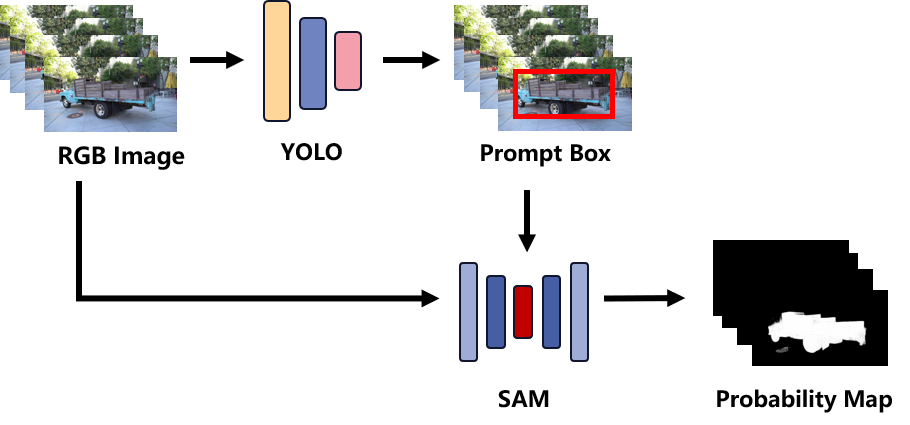}
  \caption{The pipeline of probability mask generation.}
  \label{fig:fig2}
\end{figure}

To generate the probability mask, as illustrated in Fig.~\ref{fig:fig2}, we first use a YOLO detector to process each RGB image and obtain the bounding box of the target object. This bounding box then serves as a prompt for the foundation model SAM to segment the target object from the image. Finally, the output from SAM is further refined by replacing the conventional binarization step with a mapping of the values to the [0,1] interval, resulting in a continuous probability mask. In this mask, each pixel value represents the probability of that pixel belonging to the target object.

In standard 2DGS pipleline, the sparse point cloud generated by Structure‑from‑Motion (SfM) serves as the initial point cloud for the Gaussian primitives. As a classical approach, SfM achieves sparse 3D reconstruction without requiring prior camera calibration, offering strong generalizability. However, directly using this point cloud for initialization can lead to issues such as redundant background inclusion and mis‑segmentation of target samples.

\textbf{Background-Redundant Initialization Issue.} For simplicity, we refer to the points on the object surface as the foreground point cloud, and the points not on the target object as the background point cloud. Since SfM is solely based on multi-view image feature matching and lacks effective integration of semantics, the resulting initial point cloud contains a large number of background points. Once the background points are adopted to initialize Gaussian primitives, significant redundant background Gaussians will be unnecessarily introduced in subsequent training. To address this issue, we propose a general foreground point cloud extraction method, in which the point cloud is probabilistically evaluated based on co-visibility constraints, enabling the extraction of high-confidence foreground points that are then used for Gaussian initialization.  Extracting foreground points not only effectively reduces the number of redundant Gaussian primitives but also provides a reliable basis for subsequently mitigating the sample mis-segmentation problem.

\begin{figure*}
  \centering
  \includegraphics[width=\linewidth]{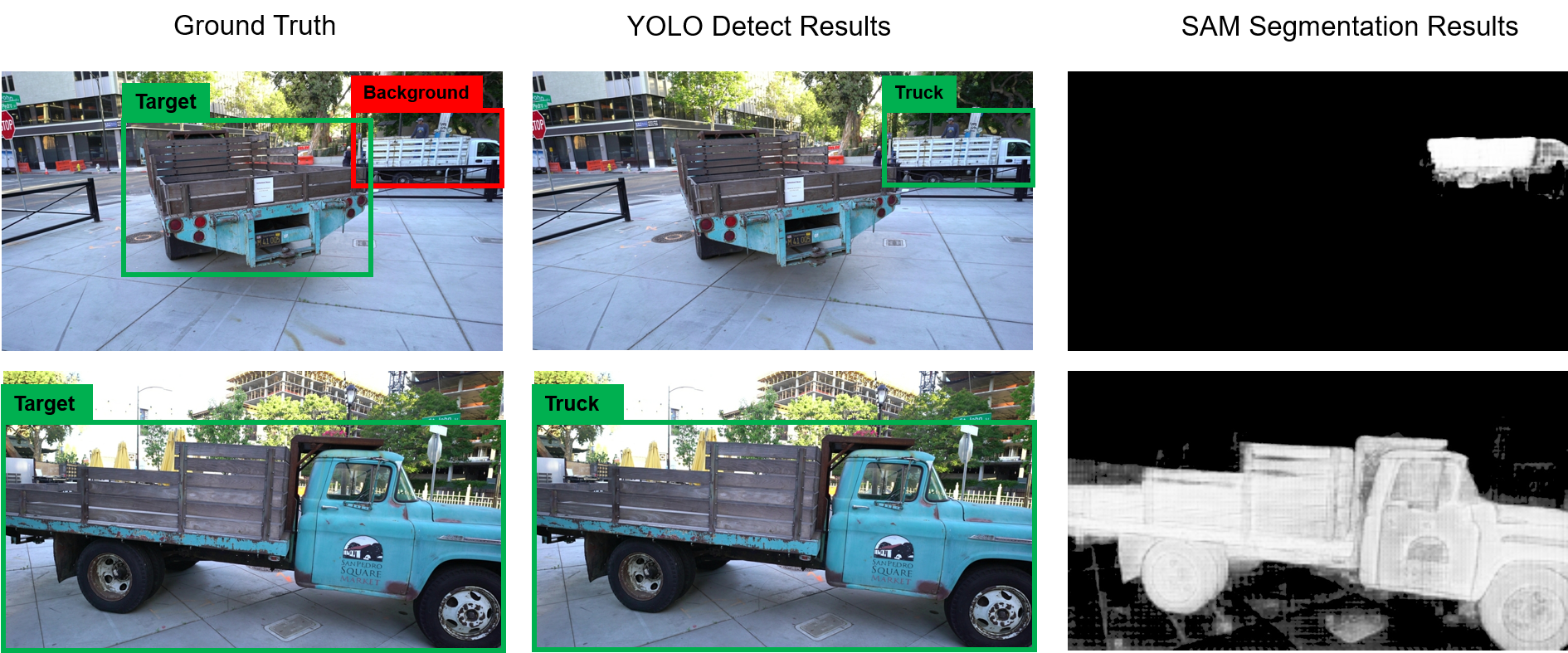}
  \caption{Examples of probability masks. The first row shows erroneous probability masks caused by target detection failures, while the second row presents probability masks with blurred boundaries. The former leads to severe optimization errors and needs to be removed from the dataset, whereas the latter can be corrected through multi-view optimization and is therefore considered a tolerable error.}
  \label{fig:fig3}
\end{figure*}

\textbf{Sample Mis-Segmentation Issue.} The generation of probability masks relies on the target candidate bounding boxes provided by YOLO. However, in complex scenes, the presence of multiple object instances within a single image can lead to erroneous identification of certain background regions as foreground. Specifically, as illustrated in Fig.~\ref{fig:fig3}, the presence of multiple truck instances within the same viewpoint causes the detection model to mistakenly classify parts of the background as target objects. Unlike segmentation errors that merely manifest as blurred or imprecise boundaries, object misidentification has a much more detrimental impact on the training process. During model optimization, both the photometric loss and the regularization loss rely on the probability masks to constrain the foreground regions. When mis-classification occurs, the probabilities of Gaussian primitives within true foreground regions may be incorrectly suppressed, meanwhile those associated with background regions are abnormally amplified. To address this issue, we systematically evaluate the quality of probabilistic segmentation results to exclude samples with severe mis-classification. Based on the foreground point cloud extracted using the aforementioned method, we further propose a low-quality sample filtering method. This method projects the foreground point cloud onto the two-dimensional image space. By analyzing the global characteristics of the projected distribution, we can distinguish high-quality probabilistic segmentation results from low-quality ones. 

\subsection{Data Refinement}
\label{sec:data refinement}

\textbf{Foreground Point Cloud Extraction.} To effectively extract the foreground point cloud, we propose a method that determines point cloud membership by leveraging multi-view projections. This method is motivated by the observation that, in multi-view data, semantic segmentation cues within the object region tend to exhibit higher consistency and confidence. Therefore, by statistically fusing multi-view semantic probabilities, we can effectively measure the foreground confidence of each point. Specifically, the initial point cloud can be represented as $G = \{ X_i \mid i = 1, \ldots, N \}$, where $X_i \in \mathbb{R}^3$ denotes the 3D coordinates of the $i$-th point, and $N$ denotes the total number of points in the point cloud. Meanwhile,camera views can be represented as $C=\{V_j \mid j = 1,\ldots, M\}$, where $V_j$ denotes the $j$-th camera view and $M$ is the total number of views. We project the sparse point cloud $G$ into all views $C$. For each point $X_i$, its visibility in view $V_j$ is determined by whether its projected image coordinates $\mathbf{x}_i^j$ fall within the valid image region. Here, $\mathbf{x}_i^j \in \mathbb{R}^2$ denotes the image coordinates of the projection of the point $X_i$ onto the image plane of view $V_j$. We then compute the mean of the probability values at the projected locations of point $X_i$ across all visible views, yielding an average probability that represents the confidence of point $X_i$ belonging to the foreground target. Based on the confidence metric, we filter the initial point cloud to obtain the foreground point set $G' = \{ X_i' \mid i = 1, \ldots, N' \}$. Here, $X_i' \in \mathbb{R}^3$ denotes the coordinates of the $i$-th filtered 3D point, and $N'$ is the total number of points after filtering, with $N' \leq N$.
 The confidence score $S(X_i)$ for point $X_i$ is computed as follows:
\begin{equation}
S( X_i ) = \frac { \sum _ { j = 1 } ^ { M } I_i^j \cdot P _ j ( \mathbf{x}_i^j ) } { \sum _ { j = 1 } ^ { M } I_i^j } , I_i^j  \in \{ 0,1 \} 
\end{equation}
where $P_j(\mathbf{x}_i^j)$ denotes the probability value at the projected location $\mathbf{x}_i^j$ under view $V_j$, and $I_i^j$ is an indicator function that denotes whether point $X_i$ within the field of view $V_j$.

\textbf{Low-Quality Samples Filtering.} To address the sample mis-segmentation issue, we propose a low-quality sample filtering strategy based on foreground point cloud projection. Specifically, we project the filtered foreground point cloud $G^\prime$ into the view $V_j$. In high-quality probability masks, the projections of these points are largely concentrated within the foreground contours, whereas in low-quality or misdetected probability masks, a substantial portion of the projected points falls into background regions. Based on this observation, we evaluate the reliability of each sample. For view $V_j$, we count the number of projected points within its field of view and compute the average probability value at the projected locations of all visible points, which reflects the overall confidence of the probability mask. Views whose confidence scores fall below a predefined threshold are discarded, and only high-confidence segmentation results are retained for subsequent training. The confidence score $S(V_j)$ for view $V_j$ is computed as follows:
\begin{equation}
S(V_j)=\frac{1}{\sum_{i=1}^{N^\prime}I_i^j}\sum_{i=1}^{N^\prime}I_i^j\cdot P_j(\mathbf{x}_i^j),I_i^j\in\{0,1\}
\end{equation}

The data refinement strategy effectively excludes background points during the initialization of 3D Gaussians. Meanwhile, by discarding low-quality samples, the model receives more accurate supervision during training, which further enhances the detail fidelity and overall stability of the reconstruction results.

\subsection{Probabilistic Single-Object Gaussian Model and Optimization}
\label{sec:Probabilistic Single-Object Gaussian Model and Optimization}

\textbf{Gaussian Probability Rendering.} By using an explicit 3D representation, 2DGS begins by initializing Gaussian primitives from a sparse point cloud and optimizes their attributes to fit the scene geometry and appearance. During rendering, 2DGS uses differentiable rasterization to accumulate the contribution of each Gaussian primitive to the image plane. Specifically, each pixel is influenced by a set of Gaussian primitives intersected by the corresponding viewing ray, and its color is blended via alpha blending as follows:
\begin{equation}
\label{Eq:c_blender}
{r}=\sum_{i=1}^{n_k}{c_i\alpha_iT_i}
\end{equation}
where $i$ indexes the Gaussian primitives sorted in a front-to-back order along the specific viewing ray, and $n_k$ denotes the total number of Gaussians intersected by that ray. $c_i$ denotes the color of the $i$-th Gaussian primitive, and $\alpha_i$ represents its blending weight, which is jointly determined by the density attribute and the opacity of the Gaussian. $T_{i}=\prod_{j=1}^{i-1}(1-\alpha_{j})$ denotes the transmittance, characterizing the accumulated contribution of light passing through the preceding $i-1$ Gaussian primitives before reaching the $i$-th one.

To model the foreground probability of each Gaussian primitive, we introduce a learnable attribute $p$ for every Gaussian. During rendering, along with the color blending in Eq.~\eqref{Eq:c_blender}, a corresponding probability value {$w$} is composited for each pixel as follows:
\begin{equation}
w=\sum_{i=1}^{n_{k}}{p_i\alpha_iT_i}
\end{equation}
where, $p_i$ denotes the probability of the $i$-th Gaussian primitive.

\textbf{Loss Function Definition.} To effectively supervise the joint optimization of probabilities and other Gaussian parameters, we design a set of corresponding loss functions for the Gaussian primitives. The overall loss is composed of two components: the first component consists of the original loss terms in the 2DGS framework, including the photometric loss and several geometric regularization terms, such as depth distortion loss and normal consistency loss; the second is a dedicated probability loss designed to supervise the probabilistic attribute. For the photometric loss, ground-truth RGB images usually contain a significant amount of background content, whereas our reconstruction task only concerns the target object; therefore, background regions are masked out during training. Specifically, we utilize the probability masks to determine whether each pixel belongs to the target object, and we weight the photometric loss using the probability mask to suppress background interference.

For the photometric loss, we adopt the formulation of the original 2DGS approach and decompose it into two terms. The first term is a per-pixel $L_1$ loss that enforces photometric consistency. The second term is an SSIM-based loss term, which evaluates the preservation of local structural information. Meanwhile, both terms are evaluated only on foreground pixels corresponding to the target object, while masking out background discrepancies. The combination of these two terms helps ensure accurate color reproduction and improves the structural fidelity of the reconstructed results. The loss is formulated as follows:
\begin{align}
\label{Eq:c_loss_5}
\mathcal{L}_1&=L_1(P_{pred} \odot R_{gt}, P_{pred} \odot R_{render});\\ 
\label{Eq:c_loss_6}
\mathcal{L}_s&=SSIM(P_{pred} \odot R_{gt}, P_{pred} \odot R_{render});\\
\mathcal{L}_c&=(1-\lambda_1)\mathcal{L}_1+\lambda_1(1 - \mathcal{L}_s).  
\end{align}

 Eq.~\eqref{Eq:c_loss_5} and Eq.~\eqref{Eq:c_loss_6} define the $L_1$ loss and the SSIM-based loss, respectively. The operator $\odot$ denotes element-wise multiplication. $P_{pred}$ denotes the probability mask predicted by SAM, while $R_{render}$ and $R_{gt}$ represent the rendered RGB image and the ground-truth RGB image, respectively. In addition, $\lambda_1$ is a balancing coefficient.

Inspired by the photometric loss design, we adopt a $L_1$ Loss to optimize the probabilistic attribute of 2D Gaussian primitives. Specifically, we compute the $L_1$ loss between  the predicted probability mask $P_{pred}$ and the rendered probability mask $P_{render}$, encouraging 2D Gaussian primitives to learn the correct foreground probabilities. The probability loss is computed as follows:
\begin{equation}
\mathcal{L}_{p}=\frac{1}{n_{pix}}\sum_{k=1}^{n_{pix}}\left|w_k-w_k'\right|
\end{equation}
where $w_k$ denotes the rendered probability of the $k$-th pixel, {$w'_k$} denotes the predicted probability of the $k$-th pixel by SAM, and $n_{pix}$ denotes the total number of pixels in the current view.

The depth distortion loss is primarily used to regularize the depth-wise distribution of  Gaussian primitives along the viewing ray, preventing degenerate cases where the related primitives are overly dispersed around the real physical surface. Specifically, we denote the depth distribution map as $D_d$, where each pixel encodes the estimated depth distribution along its viewing direction, thereby characterizing the degree of geometric dispersion and uncertainty along the ray. The normal consistency loss is primarily used to regularize the surface geometry by encouraging consistent normal directions among neighboring Gaussian primitives, thereby promoting local geometric smoothness during rendering. Specifically, the rendered depth map $D_{render}$ is used to compute spatial gradients via finite differences, yielding the local surface normals $N_{local}$. The rendered normal map $N_{render}$ is obtained through weighted aggregation of Gaussian normals along each viewing ray. By enforcing alignment between $N_{local}$ and $N_{render}$, we enhance the geometric fidelity of the reconstruction. The depth distortion and normal consistency losses are computed as follows:
\begin{align}
\mathcal{L}_{d} & =\frac{1}{n_{pix}}\sum_{k=1}^{n_{pix}}w_k'\sum_{j=1}^{n_k}\sum_{i=1}^{j-1}\omega_{i}\omega_{j}|z_{i}-z_{j}|\\
\mathcal{L}_n & =\frac{1}{n_{pix}}\sum_{k=1}^{n_{pix}}w_k'\sum_{i=1}^{n_k}\omega_i(1-\textcolor{black}{n_i n'})
\end{align}
where, $\omega_i = \alpha_i T_i$ denotes the blending weight of the $i$-th Gaussian primitive, and $z_i$ denotes its intersection depth along the viewing ray. $n_i$ denotes the normal vector of the $i$-th 2D Gaussian primitive, and $n'$ denotes the surface normal estimated from the Gaussian depth information. Specifically, $n'$ is computed using finite differences over neighboring depth samples as follows:
\begin{equation}
n'(x,y) = \frac{\nabla_x p_s \times \nabla_y p_s}{\left\lVert \nabla_x p_s \times \nabla_y p_s \right\rVert}
\end{equation}
where $(x, y)$ denotes the pixel coordinates on the image plane. $p_s$ represents the 3D surface point corresponding to pixel $(x, y)$, which is obtained by back-projecting the depth value into the camera coordinate system. $\nabla_x p_s$ and $\nabla_yp_s$ denote the partial derivatives of $p_s$ with respect to the image coordinates $x$ and $y$, respectively.

For each image observation with known camera pose and its corresponding probability mask, the overall loss function is formulated as follows, where $\alpha$, $\beta$, and $\gamma$ are balancing coefficients. In our implementation, we set 
$\alpha = 100$, $\beta = 0.8$, and $\gamma = 0.8$.
\begin{equation}
\mathcal{L}=\mathcal{L}_{c}+\alpha\mathcal{L}_{d}+\beta\mathcal{L}_{n}+\gamma\mathcal{L}_{p}
\end{equation}


\textbf{{Two-Stage Segmentation Strategy}}. To further improve reconstruction accuracy and stability in object boundary regions, we conduct a systematic analysis and refinement of the probability mask utilization strategy during training. Existing approaches typically adopt probability masks predicted by SAM as supervision signals to guide the optimization of Gaussian primitives. However, in practice, we observe that SAM often produces low-confidence transition bands near object boundaries. When such probability masks are continuously used throughout the entire training process, the optimization in boundary regions becomes unstable, which frequently leads to blurred boundaries and indistinct contours in the reconstructed results.

Further analysis reveals that these low-confidence boundary regions lack spatial consistency across different viewpoints. Although probability masks may exhibit significant uncertainty near object boundaries in individual views, the Gaussian model can still converge to stable and plausible geometric structures under multi-view constraints after sufficient training iterations. Motivated by this observation, we propose a probability mask replacement strategy that fully exploits the geometric representation capability of the model in the middle-to-late stages of training.

During the early training stage, SAM-generated probability masks are used to guide the model toward the target object and suppress background interference. After 7,000 training iterations, when the overall geometric structure has largely converged, we render probability masks for all training views using the current model. These rendered probability masks are then used to replace the original SAM-generated masks as the supervision signals in the subsequent training stage. Owing to their higher cross-view consistency, the rendered probability masks provide more stable and sharper constraints along object boundaries, thereby effectively alleviating boundary uncertainty introduced by segmentation errors.

As demonstrated by the experimental results in Sec ~\ref{sec:Ablation Study}, compared with configurations that rely solely on SAM-generated probability masks throughout training, the proposed probability mask replacement strategy significantly enhances the clarity of object boundary contours, reduces reconstruction noise near boundaries, and further improves both boundary reconstruction accuracy and overall reconstruction stability in quantitative evaluations.

\section{Experiments}
To comprehensively evaluate our approach, we conduct extensive qualitative and quantitative experiments on multiple public datasets. For qualitative evaluation, we select representative scenes from the MIP-360 \cite{Barron2021}, T\&T \cite{Knapitsch2017}, and LLFF \cite{Mildenhall2019} datasets to visually illustrate and compare reconstruction quality. For quantitative evaluation, we first analyze the model compression capability by measuring the number of Gaussians on the Truck scene from the T\&T dataset. We then evaluate the reconstruction accuracy of the proposed method on the NVOS \cite{Ren2022} dataset. Finally, by masking out background regions, we further quantify the visual reconstruction quality on object surfaces. For evaluating the contribution of individual modules, we perform ablation studies by using the Truck scene from the T\&T dataset and a Blender-rendered synthetic dataset with accurate masks.

\subsection{Qualitative Comparison}
As illustrated in Fig.~\ref{fig:fig4}, we perform a qualitative evaluation using a set of representative single-object scenes from widely used 3D reconstruction benchmark datasets to assess the effectiveness of our approach. For each scene, we select one representative view for visualization to demonstrate the quality of our single-object reconstruction and to compare the rendered probability masks with the SAM-generated probability masks used for training.

The experimental results show that our method can accurately and efficiently achieve photorealistic reconstruction of single objects. Moreover, compared with the original SAM probability masks, the probability masks rendered by our method exhibit significantly sharper object boundaries, leading to a substantial improvement in the boundary quality of SAM-based segmentation. As highlighted in the third column, the SAM-predicted probability masks present a clear transition region between the object and the background. In contrast, the corresponding regions in the rendered probability masks shown in the fourth column are noticeably sharper. These results further demonstrate the robustness and effectiveness of the proposed method in extracting object boundaries within complex scenes.

\subsection{Quantitative Comparison}

\begin{figure}
  \centering
  \includegraphics[width=\linewidth]{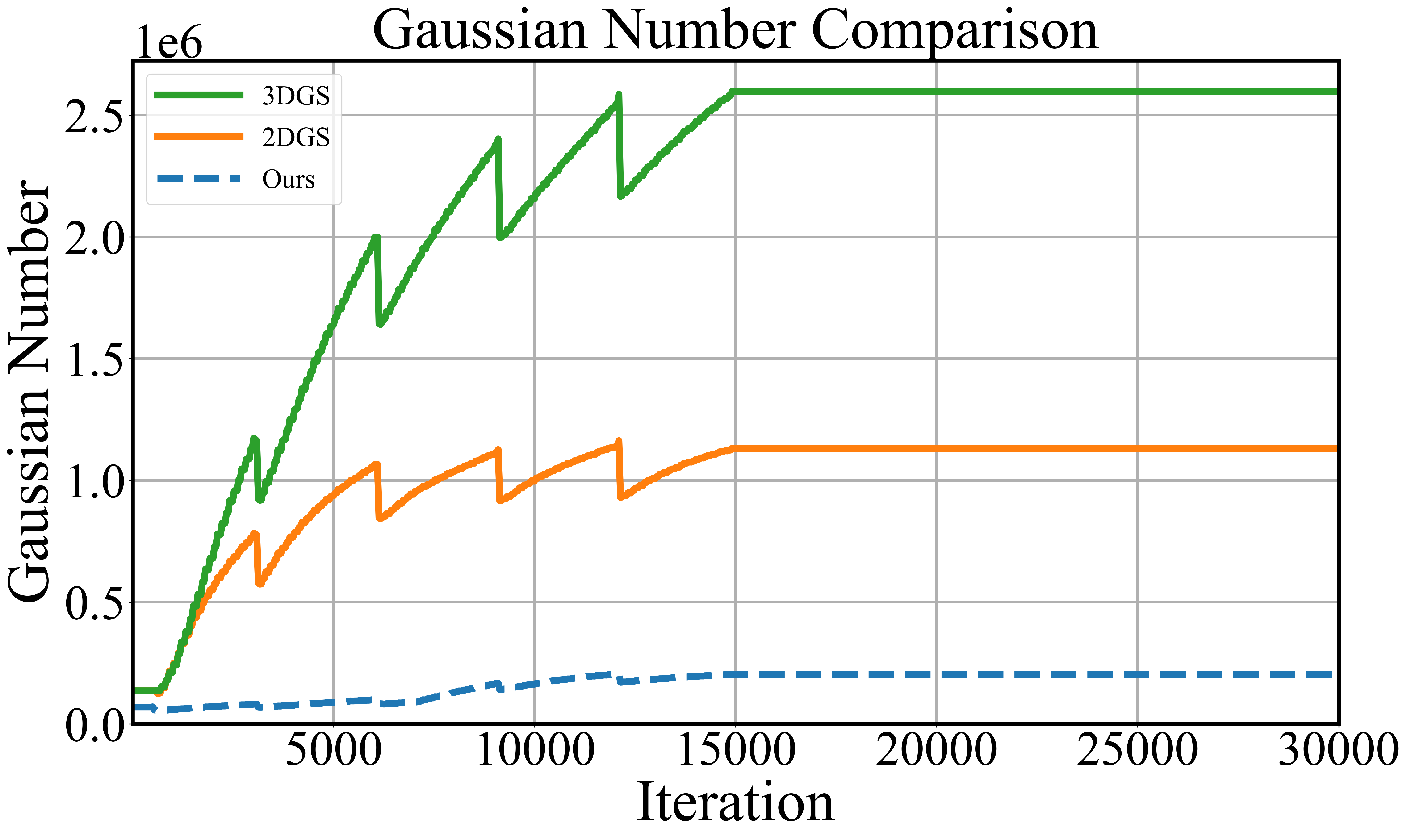}
  \caption{The growing trend in Gaussian counts during training.}
  \label{fig:fig6}
\end{figure}

\begin{figure*}[t]
  \centering
  \includegraphics[width=\linewidth]{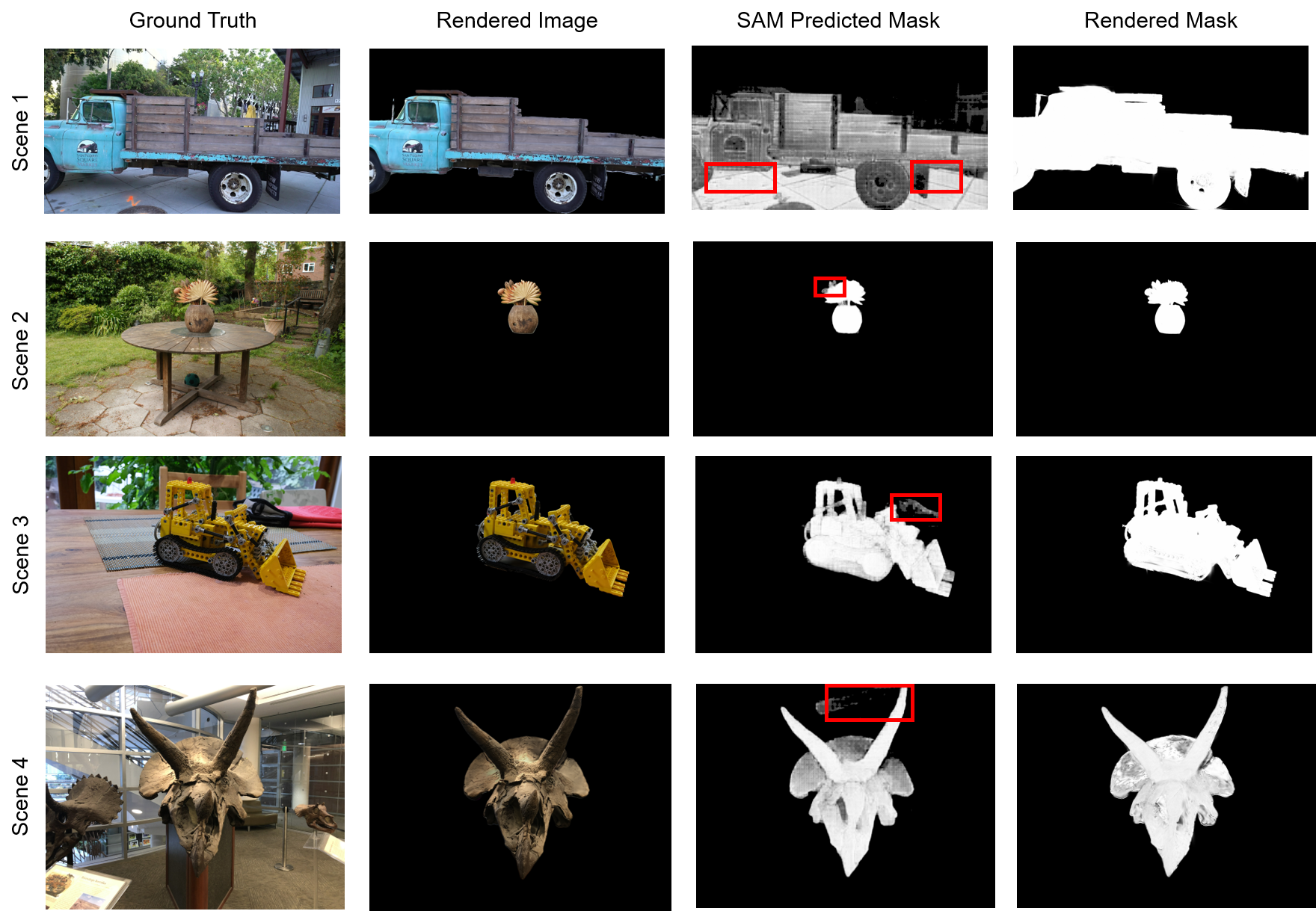}
  \caption{Qualitative comparison of our method with baseline approaches. The first row depicts the Truck scene from the T\&T dataset, the second and third rows illustrate the Garden and Kitchen scenes from the MIP-360 dataset, and the fourth row shows the Horns scene from the LLFF dataset. The first column presents the ground-truth RGB images, the second column shows the images rendered by our method, the third column displays the probability masks predicted by SAM, and the fourth column shows the probability masks rendered by our model.}
  \label{fig:fig4}
\end{figure*}

\textbf{Compression Evaluation.} We evaluate the effectiveness of our method in terms of model compression capability and memory footprint reduction. Since the number of Gaussian primitives directly reflects the model size and training resource consumption, we adopt the number of Gaussians as the primary metric for measuring memory usage. We compare our method with 3DGS and 2DGS by conducting experiments on the Truck scene using an NVIDIA RTX 3090 GPU for 30,000 training iterations, with pruning and densification disabled after iteration 15,000. {Both 3DGS and 2DGS initialize Gaussians from SfM point clouds and perform pruning based on opacity, whereas our approach first filters out the majority of background points and then conducts Gaussian pruning during training based on a combination of opacity and probability values.} As shown in Fig.~\ref{fig:fig6}, we report the number of Gaussians every ten iterations. The results demonstrate that our approach significantly reduces the number of Gaussians, achieving approximately one-fourth of that of 2DGS and nearly an order of magnitude fewer than that of 3DGS.

\textbf{Boundary Accuracy Evaluation.} We conduct a quantitative evaluation of the proposed method on the NVOS dataset. This dataset contains eight forward-facing object-centric 3D scenes, each accompanied by a ground-truth object mask for evaluation. For probability mask acquisition, we use the inference outputs of the SAM2\cite{Ravi2024} model as pixel-wise probabilistic supervision.

After training, we threshold the rendered probability masks at $\tau = 0.1$ to obtain the final object masks, and then evaluate the mean Intersection-over-Union (mIoU) and mean pixel accuracy (mAcc) over eight scenes. Meanwhile, we record the reconstruction time to evaluate the reconstruction efficiency. Since existing methods typically perform segmentation based on a fully reconstructed scene, whereas our method focuses on object-level reconstruction directly from 2D images, the reported runtime is measured starting from the scene reconstruction stage. Tab.~\ref{tab:tab1} presents a quantitative comparison of boundary accuracy for single-object reconstruction between our method and existing methods. {Among the compared methods, SAGA adopts a 3D Gaussian-based reconstruction paradigm, while the remaining methods are based on NeRF. Most of these methods follow a scene-first reconstruction pipeline, where global multi-view image features are incorporated into the model training to reconstruct the entire scene, and the target object is subsequently extracted based on prompt information, without explicitly focusing on the target region during reconstruction. The optimization over background regions not only increases the computational and optimization burden, but also does not effectively contribute to improving the reconstruction quality of the target object. In contrast, our method suppresses background interference by leveraging probability masks during reconstruction, and explicitly focuses on optimizing the Gaussian attributes associated with the target object, thereby improving reconstruction efficiency while preserving high-quality target reconstruction.}

Experimental results indicate that although our method achieves slightly lower contour accuracy than existing methods, it still maintains competitive performance. Moreover, it significantly reduces the overall runtime compared with previous approaches, further demonstrating its effectiveness and practical applicability.

\begin{table}[htbp]
  \centering
  \caption{Quantitative results on NVOS dataset. Bold numbers indicate the best results. “Full Scene” denotes whether full-scene reconstruction is required, and “Total” represents the time required from scene training to completing a single scene segmentation.}
  \label{tab:tab1}
  \small
  \setlength{\tabcolsep}{4pt}
  \begin{tabular}{lcccc}
    \toprule
    Method & Full Scene & mIOU(\%) & mAcc(\%) & Total \\
    \midrule
    NVOS\cite{Ren2022} & need & 39.4 & 73.6 & -- \\
    ISRF\cite{Goel2023} & need & 70.1 & 92.0 & $>$1 h \\
    SGISRF\cite{Tang2023} & need & 83.8 & 96.4 & -- \\
    SA3D\cite{Cen2023} & need & 90.3 & 98.2 & $>$1 h\\
    SAGA\cite{Cen2025a} & need & \textbf{90.9} & 98.3 & 3320 s\\
    \midrule
    Ours & \textbf{no need} & 90.1 & \textbf{98.4} & \textbf{1130 s} \\
    \bottomrule
  \end{tabular}
\end{table}

\textbf{Reconstruction Quality Evaluation.} To evaluate the performance of our single-object reconstruction, we conduct experiments on the Truck scene and the Kitchen scene.

\begin{figure}
  \centering
  \includegraphics[width=\linewidth]{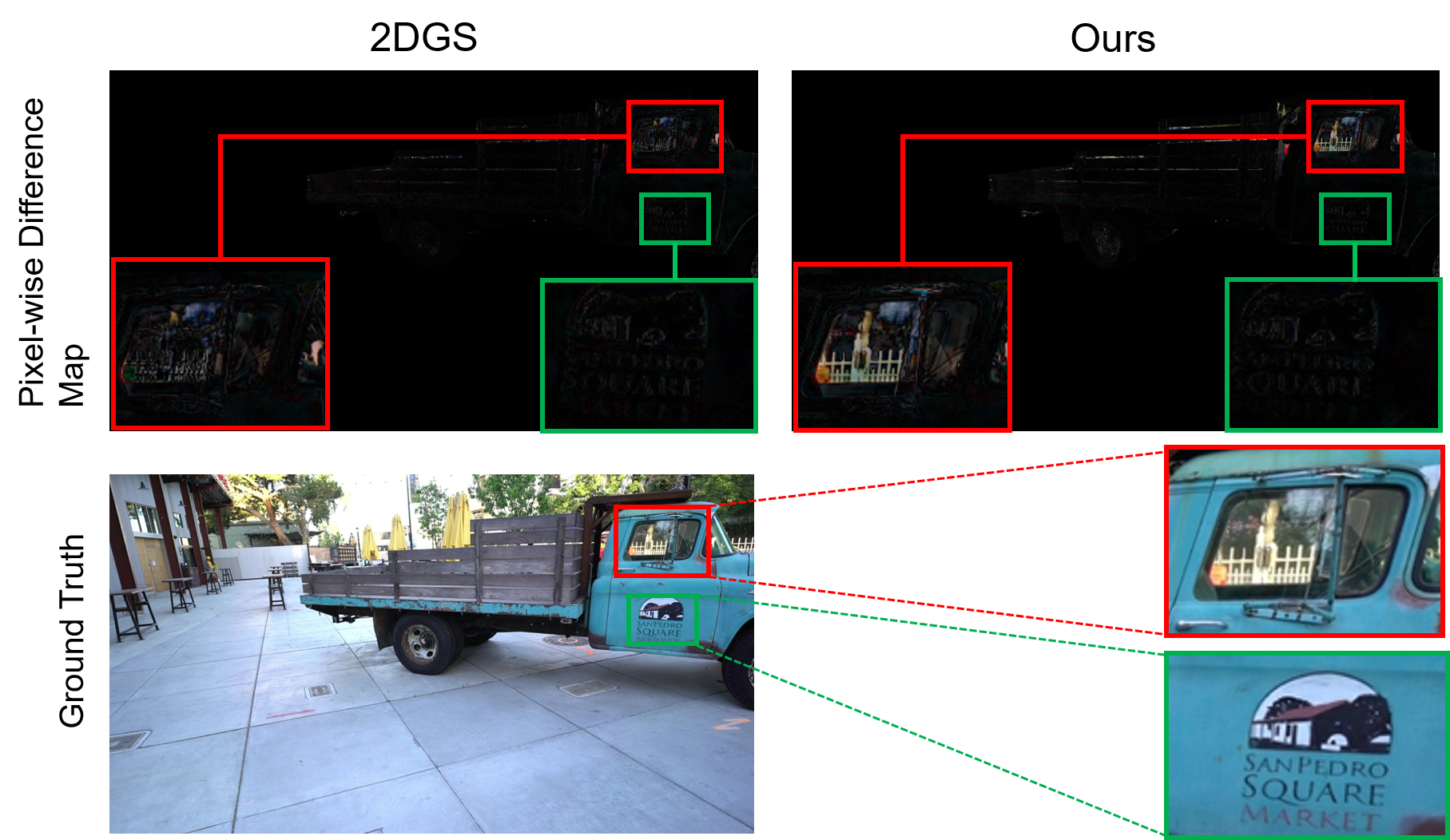}
  \caption{The difference between the rendered RGB image and the ground-truth RGB image. The first row shows the difference computed after removing background regions using SAM masks, while the second row presents the ground-truth image for comparison. Red regions indicate transparent areas near object boundaries, whereas green regions correspond to non-transparent regions on the target surface.}
  \label{fig:fig5}
\end{figure}

As shown in Tab.~\ref{tab:tab2}, although our method performs slightly below 2DGS in certain metrics, any observed differences mainly stem from limitations in segmentation accuracy rather than from deficiencies in reconstruction performance. As shown in Fig.~\ref{fig:fig5}, in the Truck scene, transparent glass regions cannot completely exclude background influence, which amplifies apparent errors, while reconstruction of opaque regions remains highly consistent with the ground-truth images. Overall, our method achieves object-level reconstruction quality comparable to 2DGS, while substantially reducing the number of Gaussians required.

\begin{table}[htbp]
  \centering
  \caption{Quantitative comparison of 2D Gaussian Splatting and our method.}
  \label{tab:tab2}
  \begin{tabular}{lcccc}
    \toprule
    \multirow{2}{*}{Method} & \multicolumn{2}{c}{Truck} & \multicolumn{2}{c}{Kitchen} \\
    \cmidrule(lr){2-3} \cmidrule(lr){4-5}
    & PSNR & SSIM & PSNR & SSIM \\
    \midrule
    2DGS\cite{Huang2024} & 31.2 & 0.96 & 37.2 & 0.98 \\
    Ours & 29.0 & 0.95 & 33.6 & 0.98 \\
    \bottomrule
  \end{tabular}
\end{table}

\subsection{Ablation Study}
\label{sec:Ablation Study}
\begin{figure*}
  \centering
  \includegraphics[width=\linewidth]{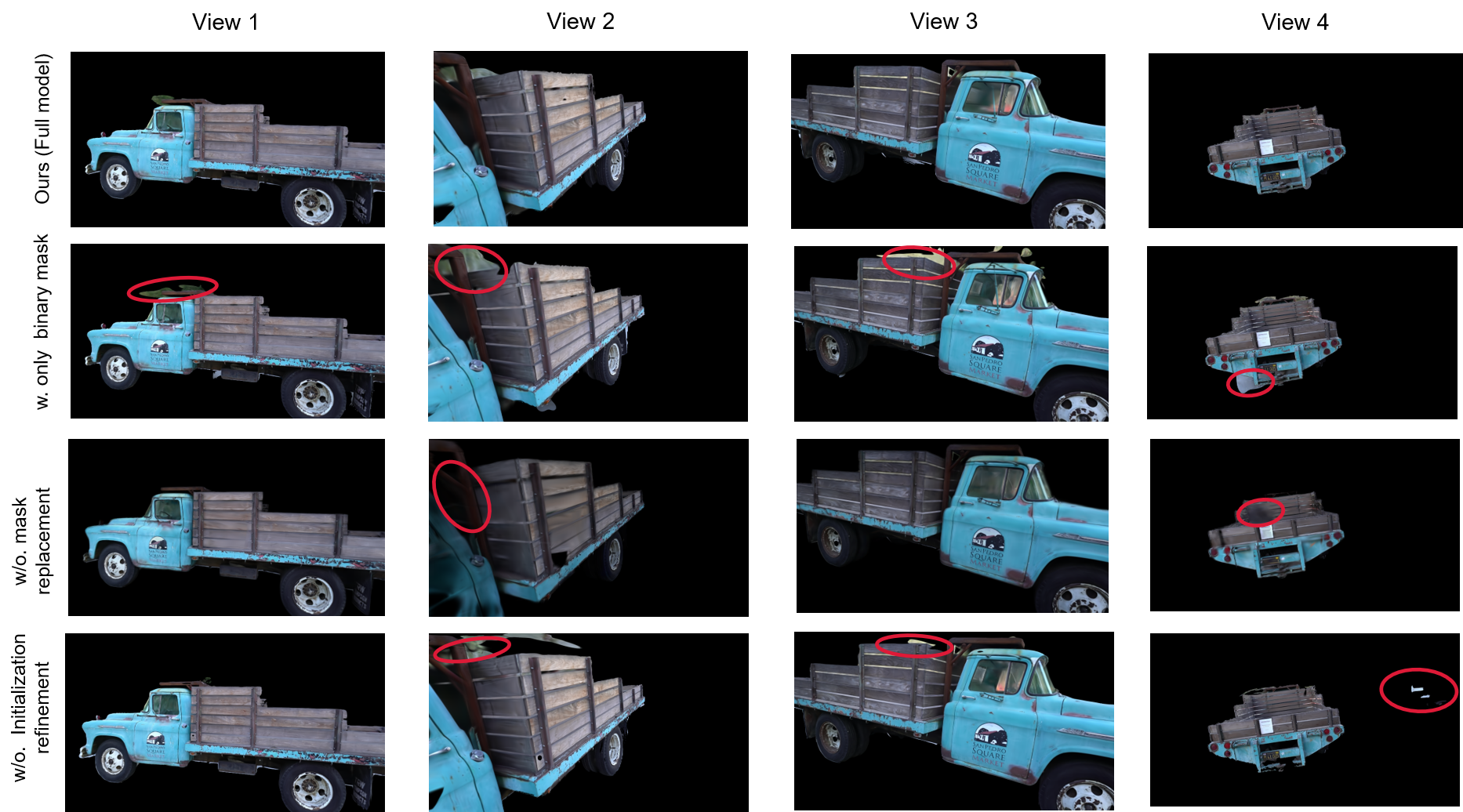}
  \caption{Qualitative comparison of ablation study results. Each column corresponds to a specific view, while the second to fourth rows modify a single factor at a time relative to the first row. The first row shows reconstruction results obtained using all three proposed strategies: probability-mask replacement, continuous probability masks, and data refinement. The second row replaces the continuous probability masks with binary (0–1) masks, while keeping all other settings unchanged. The third row removes the probability-mask replacement step and relies exclusively on SAM-generated probability masks throughout training. The fourth row removes the data refinement process and performs reconstruction directly from the original inputs.}
  \label{fig:fig7}
\end{figure*}

\begin{figure}
  \centering
  \includegraphics[width=\linewidth]{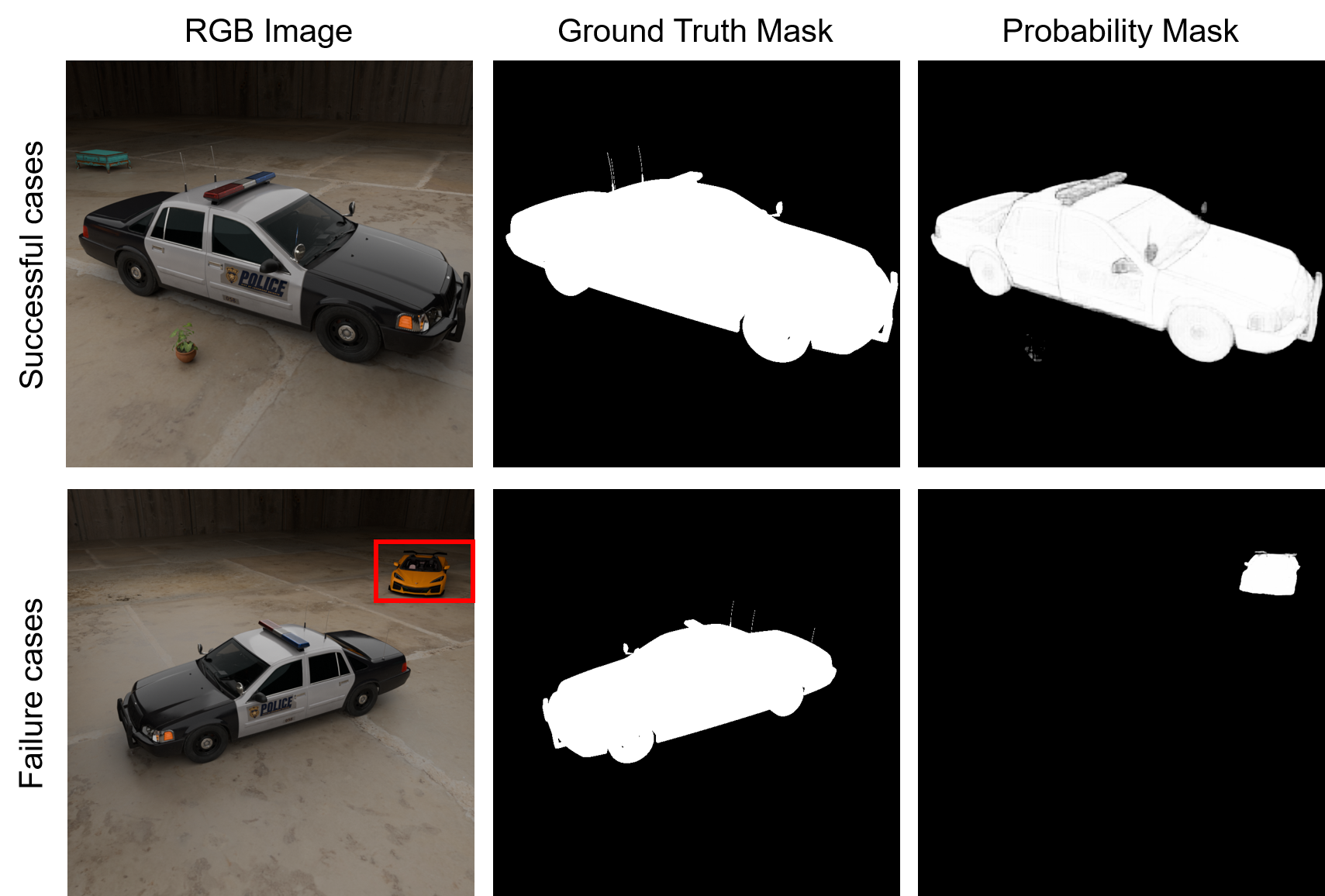}
  \caption{Examples from the Blender-rendered synthetic dataset. The first column shows the RGB images, the second column presents the ground-truth object masks rendered by Blender, and the third column displays the probability masks used for training, predicted by the SAM model. The first row illustrates successful segmentation examples, while the second row shows failure cases in which background vehicles (highlighted by red boxes) are incorrectly classified as foreground. The dataset contains 200 samples in total, approximately 10\% of which are failure cases, designed to simulate noise and inaccuracies encountered in real-world scenarios.}
  \label{fig:fig8}
\end{figure}

In the ablation study, we investigate three key aspects: the effect of probability-mask replacement on reconstruction quality, the comparative performance between binary masks and probability masks, and the impact of data refinement versus non-filtering strategies. We conduct experiments on the Truck scene from the T\&T dataset, and Fig.~\ref{fig:fig7} shows representative examples. {In this experiment, to quantitatively evaluate the effectiveness of the three strategies, we adopt mIoU and mAcc as metrics to measure the boundary accuracy of the reconstructed models. Since the computation of mIoU and mAcc requires pixel-wise accurate masks, which are not available in existing public 3D reconstruction datasets, we employ a synthetic dataset rendered by Blender along with its ground-truth masks as the evaluation benchmark, as illustrated in Fig.~\ref{fig:fig8}.} The following sections provide a detailed quantitative and qualitative analysis of each configuration, enabling a comprehensive assessment of the individual contributions of each strategy to overall single-object reconstruction performance.

\textbf{Binary Mask vs Probability Mask.} We compare the reconstruction results obtained using binary masks versus continuous probability masks. {As shown in the first and second rows of Fig.~\ref{fig:fig7}, reconstruction with binary masks introduces spurious Gaussian primitives near object boundaries, whereas training with continuous probability masks preserves significantly cleaner and sharper boundaries.} This difference can be attributed to the fact that the hard edges of binary masks amplify boundary inaccuracies, particularly for objects that are challenging for SAM to segment precisely. In contrast, continuous probability masks provide smoother and more informative boundary supervision during training, allowing the model to progressively refine its geometry and converge toward the correct object shape over multiple training iterations.

\textbf{Probability Mask Replacement vs SAM Probability Mask.} Because SAM produces blurred segmentation boundaries in some scenes, we introduce an adaptive probability-mask refinement strategy that exploits the geometric consistency of Gaussian primitives during training. Specifically, after 7,000 training iterations, we fix the model parameters and replace the probability masks generated by SAM with those rendered by the Gaussian model. As shown in the third row of Fig.~\ref{fig:fig7}, when SAM is used throughout training, additional background Gaussians emerge on the top of the truck, and the cargo bed becomes visibly blurred. Moreover, because the probability masks generated by SAM contain excessively high-confidence values, many black Gaussian artifacts appear near structural joint regions. The underlying issue is that, in several frontal-view images, the interior of the cargo bed is incorrectly classified as background, causing its appearance to be erroneously optimized toward black during RGB reconstruction. In contrast, after applying probability-mask replacement, the updated probability masks more accurately reflect the current model geometry, thereby preventing such mis-classification in subsequent training stages. Overall, this comparison demonstrates that the probability-mask replacement strategy yields sharper object boundaries and cleaner background regions.

\textbf{Data Refinement vs Non-Refinement.} We further evaluate the impact of the data preprocessing strategy on reconstruction quality. As shown in the fourth row of Fig.~\ref{fig:fig7}, a large number of isolated Gaussians appear in distant regions. This issue arises because open scenes often contain multiple instances of the same object category, causing YOLO to mistakenly identify background regions as the target object. To address this problem, we adopt the combined point-cloud filtering and view-filtering strategy described in Sec ~\ref{sec:data refinement}. Experimental results demonstrate that this preprocessing step not only reduces computational overhead but also produces a significantly cleaner background, thereby improving the overall stability and reliability of the reconstruction.

{\textbf{Quantitative Evaluation.} To quantitatively analyze the impact of the three strategies on reconstruction accuracy, we conduct comparative experiments on the constructed synthetic dataset by progressively enabling each component. The dataset consists of 200 samples, among which five are randomly selected as the test set. Each configuration is trained for 30,000 iterations. The quantitative results are reported in Tab.~\ref{tab:tab3}. It can be observed that, compared with binary masks, probability masks achieve a substantial improvement in boundary accuracy, confirming their superior ability to supervise the optimization of Gaussian primitives. In addition, both the probability mask replacement strategy and the data refinement strategy further enhance the boundary accuracy to varying degrees, demonstrating the effectiveness of these two components. It is worth noting that, since mAcc does not penalize over-coverage, the configuration that uses only probability mask can still obtain a relatively high mAcc despite exhibiting excessive coverage, whereas its mIoU remains low, indicating inferior region-level consistency.}

\begin{table}[htbp]
    \centering
    \caption{Ablation study results on the synthetic dataset. “PM” indicates probability mask; “PMR” indicates probability mask replacement after 7,000 iterations; “DR” indicates data refinement; “BM” indicates binary mask.} 
    \label{tab:tab3}
    \begin{tabular}{cccc|cc}
        \toprule
        \multicolumn{4}{c|}{Component} & \multicolumn{2}{c}{Performance} \\
        \midrule
        PM & PMR & DR & BM & mIoU (\%) & mAcc (\%) \\
        \midrule
         &  &  &\checkmark & 56.71 & 99.36 \\
        \checkmark &  &  & & 84.33 & \textbf{99.75} \\
        \checkmark & \checkmark &  &  & 93.47 & 99.13 \\
        \checkmark & \checkmark & \checkmark &  & \textbf{95.63} & 99.57 \\
        \bottomrule
    \end{tabular}
\end{table}

\subsection{Application}

\begin{figure}
  \centering
  \includegraphics[width=\linewidth]{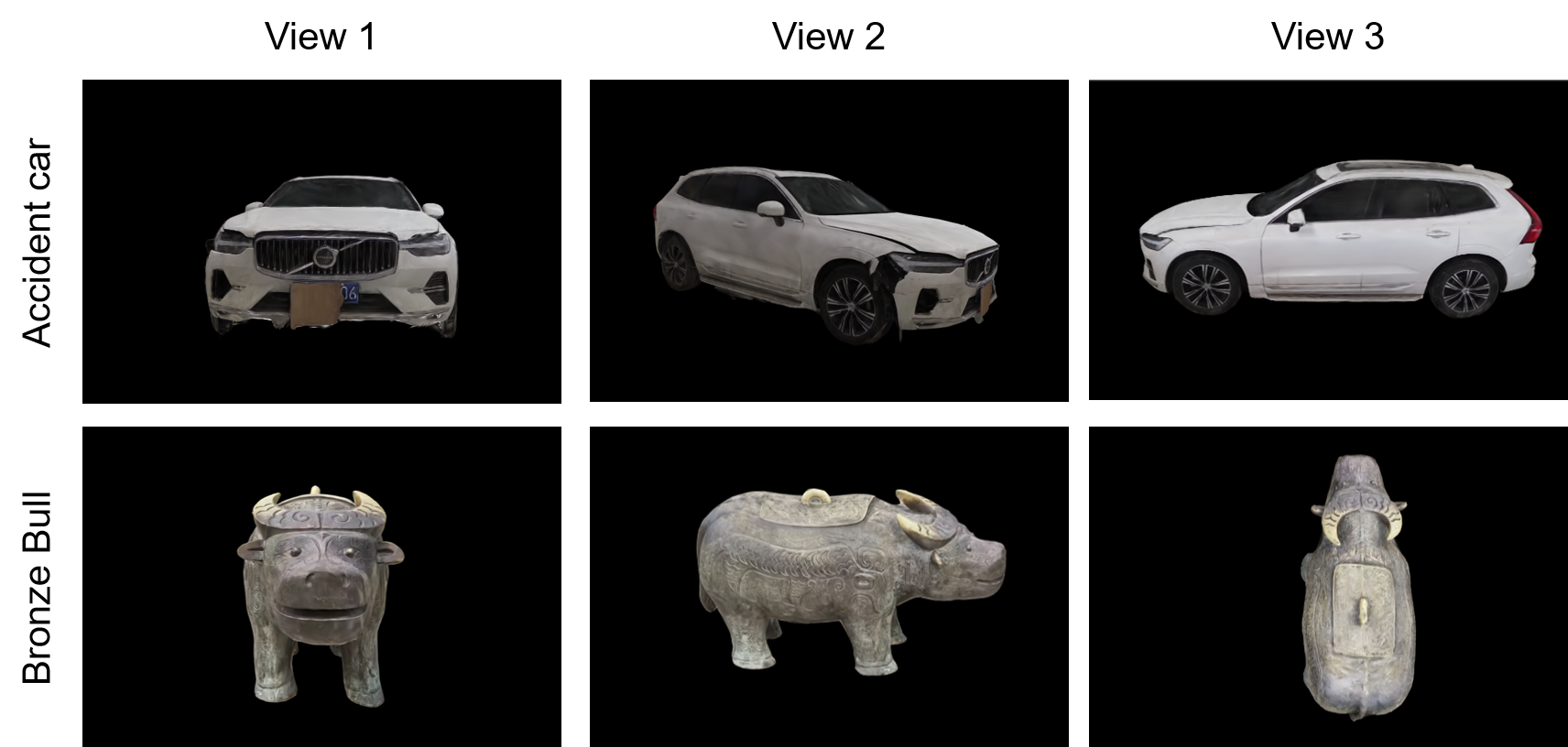}
  \caption{Application results. The first row shows the reconstruction results of the bronze artifact obtained by our method, while the second row presents the reconstruction results of the accident vehicle. Each column corresponds to the rendering results from different views.}
  \label{fig:fig9}
\end{figure}

\textcolor{black}{To evaluate the practical applicability of the proposed method, we conduct experiments in the domains of cultural heritage preservation and transportation. We select a bronze artifact and an accident vehicle as representative application scenarios, as shown in Fig.~\ref{fig:fig9}. The results demonstrate that the geometric structures and texture details on the surface of the bronze artifact are well preserved under different viewpoints. For the accident vehicle, the reconstruction faithfully captures the damaged regions, achieving high overall reconstruction accuracy and effectively avoiding the computational overhead caused by excessive background information, thereby providing reliable data support for subsequent accident analysis and investigation.}

Furthermore, to assess reconstruction efficiency, we conduct comparative experiments with the baseline 2DGS method under identical experimental settings, and analyze both reconstruction time and the number of Gaussians. As reported in Tab.~\ref{tab:tab4}, in the evaluated scenarios, the average reconstruction time of our method is about 80\% of that of 2DGS, while the required number of Gaussians is only about one-fifth of that used by 2DGS. Compared with traditional full-scene reconstruction approaches, our method significantly reduces computational resource requirements while maintaining reconstruction accuracy, thereby improving overall reconstruction efficiency and demonstrating strong potential for real-world applications.

\begin{table}[t]
  \centering
  \caption{Comparison of Gaussian Count and Training Time on Different Scenes.}
  \label{tab:tab4}
  \footnotesize
  \begin{tabular}{lcccc}
    \toprule
    \multirow{2}{*}{Method} &
    \multicolumn{2}{c}{Accident Vehicle} &
    \multicolumn{2}{c}{Bronze Artifact} \\
    \cmidrule(lr){2-3} \cmidrule(lr){4-5}
     & Time & Number 
     & Time & Number \\
    \midrule
    2DGS~\cite{Huang2024} 
      & 27.6min & 12.4$\times 10^5$ & 30.3min & 12.6$\times 10^5$ \\
    Ours 
      & \textbf{25.3min} & \textbf{2.4$\times 10^5$} & \textbf{22.0min} & \textbf{2.6$\times 10^5$} \\
    \midrule
    Reduction (\%) 
      & 8.3 & 80.6 & 27.4 & 79.4 \\
    \bottomrule
  \end{tabular}
\end{table}

\section{Conclusion and Limitations}
In this work, we propose an efficient, direct object-level 3D reconstruction framework. Our approach embeds probabilistic attributes into Gaussian primitives and dynamically removes background Gaussians during training, effectively eliminating redundant background information while preserving the primitives belonging to the target object. We leverage probability masks generated by YOLO and SAM, and further employ rendered masks to improve the learning of object boundaries. Additionally, we introduce a data refinement strategy that significantly reduces the number of initial Gaussians, thereby lowering memory consumption and computational overhead. Experimental results on multiple public datasets show that our method achieves object-level reconstruction quality comparable to existing 2DGS-based approaches, while using only about one-tenth of the Gaussians required by conventional 3DGS. This demonstrates clear advantages in model compression and computational efficiency.

Nevertheless, our method faces challenges when reconstructing objects with reflective or transparent surfaces and remains partially dependent on the segmentation accuracy of the mask generator. Future work may explore the integration of geometric constraints at object boundaries to further improve the reconstruction of objects with complex surfaces and fine structures. Overall, this work presents a lightweight and robust framework for efficient single-object 3D reconstruction, demonstrating considerable potential for practical applications.

{\small
\bibliographystyle{ieeenat_fullname}
\bibliography{ref/ref}

@Article{Kerbl2023,
  author  = {Kerbl, Bernhard and Kopanas, Georgios and Leimk{\"u}hler, Thomas and Drettakis, George},
  journal = {ACM Trans. Graph.},
  title   = {3D Gaussian splatting for real-time radiance field rendering.},
  year    = {2023},
  number  = {4},
  pages   = {139--1},
  volume  = {42},
}

@Article{Fei2024,
  author        = {Ben Fei and Jingyi Xu and Rui Zhang and Qingyuan Zhou and Weidong Yang and Ying He},
  journal       = {CoRR},
  title         = {3D Gaussian as a New Vision Era: {A} Survey},
  year          = {2024},
  volume        = {abs/2402.07181},
  archiveprefix = {arXiv},
  bibsource     = {dblp computer science bibliography, https://dblp.org},
  doi           = {10.48550/ARXIV.2402.07181},
  eprint        = {2402.07181},
}

@Article{Navaneet2023,
  author  = {Navaneet, KL and Meibodi, Kossar Pourahmadi and Koohpayegani, Soroush Abbasi and Pirsiavash, Hamed},
  journal = {arXiv preprint arXiv:2311.18159},
  title   = {Compact3d: Smaller and faster gaussian splatting with vector quantization},
  year    = {2023},
  volume  = {1},
}

@Article{Fan2024,
  author  = {Fan, Zhiwen and Wang, Kevin and Wen, Kairun and Zhu, Zehao and Xu, Dejia and Wang, Zhangyang and others},
  journal = {Advances in neural information processing systems},
  title   = {Lightgaussian: Unbounded 3d gaussian compression with 15x reduction and 200+ fps},
  year    = {2024},
  pages   = {140138--140158},
  volume  = {37},
}

@InProceedings{Lee2024,
  author    = {Lee, Joo Chan and Rho, Daniel and Sun, Xiangyu and Ko, Jong Hwan and Park, Eunbyung},
  booktitle = {Proceedings of the IEEE/CVF Conference on Computer Vision and Pattern Recognition},
  title     = {Compact 3d gaussian representation for radiance field},
  year      = {2024},
  pages     = {21719--21728},
}

@InProceedings{Lu2024,
  author    = {Lu, Tao and Yu, Mulin and Xu, Linning and Xiangli, Yuanbo and Wang, Limin and Lin, Dahua and Dai, Bo},
  booktitle = {Proceedings of the IEEE/CVF Conference on Computer Vision and Pattern Recognition},
  title     = {Scaffold-gs: Structured 3d gaussians for view-adaptive rendering},
  year      = {2024},
  pages     = {20654--20664},
}

@Article{Wang2024,
  author  = {Wang, Yufei and Li, Zhihao and Guo, Lanqing and Yang, Wenhan and Kot, Alex and Wen, Bihan},
  journal = {Advances in neural information processing systems},
  title   = {Contextgs: Compact 3d gaussian splatting with anchor level context model},
  year    = {2024},
  pages   = {51532--51551},
  volume  = {37},
}

@InProceedings{Niedermayr2024,
  author    = {Niedermayr, Simon and Stumpfegger, Josef and Westermann, R{\"u}diger},
  booktitle = {Proceedings of the IEEE/CVF Conference on Computer Vision and Pattern Recognition},
  title     = {Compressed 3d gaussian splatting for accelerated novel view synthesis},
  year      = {2024},
  pages     = {10349--10358},
}

@Article{Papantonakis2024,
  author    = {Papantonakis, Panagiotis and Kopanas, Georgios and Kerbl, Bernhard and Lanvin, Alexandre and Drettakis, George},
  journal   = {Proceedings of the ACM on Computer Graphics and Interactive Techniques},
  title     = {Reducing the memory footprint of 3d gaussian splatting},
  year      = {2024},
  number    = {1},
  pages     = {1--17},
  volume    = {7},
  publisher = {ACM New York, NY, USA},
}

@InProceedings{Cheng2024,
  author    = {Cheng, Kai and Long, Xiaoxiao and Yang, Kaizhi and Yao, Yao and Yin, Wei and Ma, Yuexin and Wang, Wenping and Chen, Xuejin},
  booktitle = {Forty-first International Conference on Machine Learning},
  title     = {Gaussianpro: 3d gaussian splatting with progressive propagation},
  year      = {2024},
}

@Article{Liu2024,
  author  = {Liu, Rong and Xu, Rui and Hu, Yue and Chen, Meida and Feng, Andrew},
  journal = {arXiv preprint arXiv:2405.12369},
  title   = {Atomgs: Atomizing gaussian splatting for high-fidelity radiance field},
  year    = {2024},
}

@InProceedings{Mallick2024,
  author    = {Mallick, Saswat Subhajyoti and Goel, Rahul and Kerbl, Bernhard and Steinberger, Markus and Carrasco, Francisco Vicente and De La Torre, Fernando},
  booktitle = {SIGGRAPH Asia 2024 Conference Papers},
  title     = {Taming 3dgs: High-quality radiance fields with limited resources},
  year      = {2024},
  pages     = {1--11},
}

@InProceedings{Kim2024,
  author    = {Kim, Sieun and Lee, Kyungjin and Lee, Youngki},
  booktitle = {Proceedings of the IEEE/CVF Conference on Computer Vision and Pattern Recognition},
  title     = {Color-cued efficient densification method for 3d gaussian splatting},
  year      = {2024},
  pages     = {775--783},
}

@Article{Chen2024,
  author    = {Chen, Danpeng and Li, Hai and Ye, Weicai and Wang, Yifan and Xie, Weijian and Zhai, Shangjin and Wang, Nan and Liu, Haomin and Bao, Hujun and Zhang, Guofeng},
  journal   = {IEEE Transactions on Visualization and Computer Graphics},
  title     = {Pgsr: Planar-based gaussian splatting for efficient and high-fidelity surface reconstruction},
  year      = {2024},
  publisher = {IEEE},
}

@InProceedings{Jiang2024,
  author    = {Jiang, Yingwenqi and Tu, Jiadong and Liu, Yuan and Gao, Xifeng and Long, Xiaoxiao and Wang, Wenping and Ma, Yuexin},
  booktitle = {Proceedings of the IEEE/CVF Conference on Computer Vision and Pattern Recognition},
  title     = {Gaussianshader: 3d gaussian splatting with shading functions for reflective surfaces},
  year      = {2024},
  pages     = {5322--5332},
}

@Article{Zhang2024,
  author  = {Zhang, Ziyu and Huang, Binbin and Jiang, Hanqing and Zhou, Liyang and Xiang, Xiaojun and Shen, Shunhan},
  journal = {arXiv preprint arXiv:2411.16392},
  title   = {Quadratic Gaussian Splatting for Efficient and Detailed Surface Reconstruction},
  year    = {2024},
}

@InProceedings{Huang2024,
  author    = {Huang, Binbin and Yu, Zehao and Chen, Anpei and Geiger, Andreas and Gao, Shenghua},
  booktitle = {ACM SIGGRAPH 2024 conference papers},
  title     = {2d gaussian splatting for geometrically accurate radiance fields},
  year      = {2024},
  pages     = {1--11},
}

@Article{Mildenhall2021,
  author    = {Mildenhall, Ben and Srinivasan, Pratul P and Tancik, Matthew and Barron, Jonathan T and Ramamoorthi, Ravi and Ng, Ren},
  journal   = {Communications of the ACM},
  title     = {Nerf: Representing scenes as neural radiance fields for view synthesis},
  year      = {2021},
  number    = {1},
  pages     = {99--106},
  volume    = {65},
  publisher = {ACM New York, NY, USA},
}

@Article{Cen2023,
  author  = {Cen, Jiazhong and Zhou, Zanwei and Fang, Jiemin and Shen, Wei and Xie, Lingxi and Jiang, Dongsheng and Zhang, Xiaopeng and Tian, Qi and others},
  journal = {Advances in Neural Information Processing Systems},
  title   = {Segment anything in 3d with nerfs},
  year    = {2023},
  pages   = {25971--25990},
  volume  = {36},
}

@Article{Chen2023,
  author  = {Chen, Xiaokang and Tang, Jiaxiang and Wan, Diwen and Wang, Jingbo and Zeng, Gang},
  journal = {arXiv preprint arXiv:2305.16233},
  title   = {Interactive segment anything nerf with feature imitation},
  year    = {2023},
}

@Article{Wang2022,
  author  = {Wang, Bing and Chen, Lu and Yang, Bo},
  journal = {arXiv preprint arXiv:2208.07227},
  title   = {Dm-nerf: 3d scene geometry decomposition and manipulation from 2d images},
  year    = {2022},
}

@InProceedings{Fu2022,
  author       = {Fu, Xiao and Zhang, Shangzhan and Chen, Tianrun and Lu, Yichong and Zhu, Lanyun and Zhou, Xiaowei and Geiger, Andreas and Liao, Yiyi},
  booktitle    = {2022 International Conference on 3D Vision (3DV)},
  title        = {Panoptic nerf: 3d-to-2d label transfer for panoptic urban scene segmentation},
  year         = {2022},
  organization = {IEEE},
  pages        = {1--11},
}

@InProceedings{Lan2024,
  author       = {Lan, Kun and Li, Haoran and Shi, Haolin and Wu, Wenjun and Wang, Lin and Liao, Yong},
  booktitle    = {2024 Asian Conference on Communication and Networks (ASIANComNet)},
  title        = {2d-guided 3d gaussian segmentation},
  year         = {2024},
  organization = {IEEE},
  pages        = {1--5},
}

@Article{Guo2024,
  author  = {Guo, Jun and Ma, Xiaojian and Fan, Yue and Liu, Huaping and Li, Qing},
  journal = {arXiv preprint arXiv:2403.15624},
  title   = {Semantic gaussians: Open-vocabulary scene understanding with 3d gaussian splatting},
  year    = {2024},
}

@InProceedings{Choi2024,
  author       = {Choi, Seokhun and Song, Hyeonseop and Kim, Jaechul and Kim, Taehyeong and Do, Hoseok},
  booktitle    = {European Conference on Computer Vision},
  title        = {Click-gaussian: Interactive segmentation to any 3d gaussians},
  year         = {2024},
  organization = {Springer},
  pages        = {289--305},
}

@InProceedings{Cen2025a,
  author    = {Cen, Jiazhong and Fang, Jiemin and Yang, Chen and Xie, Lingxi and Zhang, Xiaopeng and Shen, Wei and Tian, Qi},
  booktitle = {Proceedings of the AAAI Conference on Artificial Intelligence},
  title     = {Segment any 3d gaussians},
  year      = {2025},
  number    = {2},
  pages     = {1971--1979},
  volume    = {39},
}

@InProceedings{Ye2024,
  author       = {Ye, Mingqiao and Danelljan, Martin and Yu, Fisher and Ke, Lei},
  booktitle    = {European conference on computer vision},
  title        = {Gaussian grouping: Segment and edit anything in 3d scenes},
  year         = {2024},
  organization = {Springer},
  pages        = {162--179},
}

@Article{Hu2024,
  author  = {Hu, Xu and Wang, Yuxi and Fan, Lue and Fan, Junsong and Peng, Junran and Lei, Zhen and Li, Qing and Zhang, Zhaoxiang},
  journal = {arXiv preprint arXiv:2401.17857},
  title   = {SAGD: Boundary-enhanced segment anything in 3D Gaussian via Gaussian decomposition},
  year    = {2024},
}

@InProceedings{Shen2024,
  author       = {Shen, Qiuhong and Yang, Xingyi and Wang, Xinchao},
  booktitle    = {European Conference on Computer Vision},
  title        = {Flashsplat: 2d to 3d gaussian splatting segmentation solved optimally},
  year         = {2024},
  organization = {Springer},
  pages        = {456--472},
}

@InProceedings{Zhou2024,
  author    = {Zhou, Shijie and Chang, Haoran and Jiang, Sicheng and Fan, Zhiwen and Zhu, Zehao and Xu, Dejia and Chari, Pradyumna and You, Suya and Wang, Zhangyang and Kadambi, Achuta},
  booktitle = {Proceedings of the IEEE/CVF Conference on Computer Vision and Pattern Recognition},
  title     = {Feature 3dgs: Supercharging 3d gaussian splatting to enable distilled feature fields},
  year      = {2024},
  pages     = {21676--21685},
}

@Article{Silva2024,
  author  = {Silva, Myrna C and Dahaghin, Mahtab and Toso, Matteo and Del Bue, Alessio},
  journal = {arXiv preprint arXiv:2404.12784},
  title   = {Contrastive gaussian clustering: Weakly supervised 3d scene segmentation},
  year    = {2024},
}

@InProceedings{Zhang2025,
  author    = {Zhang, Jiaxin and Jiang, Junjun and Chen, Youyu and Jiang, Kui and Liu, Xianming},
  booktitle = {Proceedings of the Computer Vision and Pattern Recognition Conference},
  title     = {Cob-gs: Clear object boundaries in 3dgs segmentation based on boundary-adaptive gaussian splitting},
  year      = {2025},
  pages     = {19335--19344},
}

@InProceedings{Varghese2024,
  author    = {Varghese, Rejin and M., Sambath},
  booktitle = {2024 International Conference on Advances in Data Engineering and Intelligent Computing Systems (ADICS)},
  title     = {YOLOv8: A Novel Object Detection Algorithm with Enhanced Performance and Robustness},
  year      = {2024},
  pages     = {1-6},
  doi       = {10.1109/ADICS58448.2024.10533619},
}

@InProceedings{Kirillov2023,
  author    = {Kirillov, Alexander and Mintun, Eric and Ravi, Nikhila and Mao, Hanzi and Rolland, Chloe and Gustafson, Laura and Xiao, Tete and Whitehead, Spencer and Berg, Alexander C and Lo, Wan-Yen and others},
  booktitle = {Proceedings of the IEEE/CVF international conference on computer vision},
  title     = {Segment anything},
  year      = {2023},
  pages     = {4015--4026},
}

@Article{Ravi2024,
  author  = {Ravi, Nikhila and Gabeur, Valentin and Hu, Yuan-Ting and Hu, Ronghang and Ryali, Chaitanya and Ma, Tengyu and Khedr, Haitham and R{\"a}dle, Roman and Rolland, Chloe and Gustafson, Laura and others},
  journal = {arXiv preprint arXiv:2408.00714},
  title   = {Sam 2: Segment anything in images and videos},
  year    = {2024},
}

@InProceedings{Barron2021,
  author    = {Barron, Jonathan T and Mildenhall, Ben and Tancik, Matthew and Hedman, Peter and Martin-Brualla, Ricardo and Srinivasan, Pratul P},
  booktitle = {Proceedings of the IEEE/CVF international conference on computer vision},
  title     = {Mip-nerf: A multiscale representation for anti-aliasing neural radiance fields},
  year      = {2021},
  pages     = {5855--5864},
}

@Article{Knapitsch2017,
  author    = {Knapitsch, Arno and Park, Jaesik and Zhou, Qian-Yi and Koltun, Vladlen},
  journal   = {ACM Transactions on Graphics (ToG)},
  title     = {Tanks and temples: Benchmarking large-scale scene reconstruction},
  year      = {2017},
  number    = {4},
  pages     = {1--13},
  volume    = {36},
  publisher = {ACM New York, NY, USA},
}

@Article{Mildenhall2019,
  author    = {Mildenhall, Ben and Srinivasan, Pratul P and Ortiz-Cayon, Rodrigo and Kalantari, Nima Khademi and Ramamoorthi, Ravi and Ng, Ren and Kar, Abhishek},
  journal   = {ACM Transactions on Graphics (ToG)},
  title     = {Local light field fusion: Practical view synthesis with prescriptive sampling guidelines},
  year      = {2019},
  number    = {4},
  pages     = {1--14},
  volume    = {38},
  publisher = {ACM New York, NY, USA},
}

@InProceedings{Ren2022,
  author    = {Ren, Zhongzheng and Agarwala, Aseem and Russell, Bryan and Schwing, Alexander G and Wang, Oliver},
  booktitle = {Proceedings of the IEEE/CVF conference on computer vision and pattern recognition},
  title     = {Neural volumetric object selection},
  year      = {2022},
  pages     = {6133--6142},
}

@InProceedings{Goel2023,
  author    = {Goel, Rahul and Sirikonda, Dhawal and Saini, Saurabh and Narayanan, PJ},
  booktitle = {Proceedings of the IEEE/CVF conference on computer vision and pattern recognition},
  title     = {Interactive segmentation of radiance fields},
  year      = {2023},
  pages     = {4201--4211},
}

@InProceedings{Tang2023,
  author    = {Tang, Songlin and Pei, Wenjie and Tao, Xin and Jia, Tanghui and Lu, Guangming and Tai, Yu-Wing},
  booktitle = {Proceedings of the 31st ACM International Conference on Multimedia},
  title     = {Scene-generalizable interactive segmentation of radiance fields},
  year      = {2023},
  pages     = {6744--6755},
}

@InProceedings{sugar,
  author    = {Guedon, Antoine and Lepetit, Vincent},
  booktitle = {2024 IEEE/CVF Conference on Computer Vision and Pattern Recognition (CVPR)},
  title     = {{SuGaR: Surface-Aligned Gaussian Splatting for Efficient 3D Mesh Reconstruction and High-Quality Mesh Rendering}},
  year      = {2024},
  pages     = {5354-5363},
}
}

\end{document}